\documentclass{article} \usepackage{agi-10}

\nocopyright
\usepackage{latexsym,amsmath,amstext,amsfonts,amsbsy,amssymb,graphicx,hyperref}
\usepackage{algorithm,algorithmic,prettyref}
\usepackage{times} 

\newcommand{\fancy}[1]{\mathcal{#1}}

\newcommand{\algo}[1]{\textsc{#1}} 
\newcommand{\code}[1]{\begin{small}\texttt{#1}\end{small}}
\newcommand{\var}[1]{\textsf{#1}}
\newcommand{\nt}[1]{\begin{small}\textsf{#1}\end{small}}

\newrefformat{alg}{Algorithm~\ref{#1}}
\newrefformat{eq}{Equation~\ref{#1}}
\newrefformat{lem}{Lemma~\ref{#1}}
\newrefformat{def}{Definition~\ref{#1}}
\newrefformat{cor}{Corollary~\ref{#1}}
\newrefformat{chp}{Chapter~\ref{#1}}
\newrefformat{sec}{Section~\ref{#1}}
\newrefformat{apx}{Appendix~\ref{#1}}
\newrefformat{tab}{Table~\ref{#1}} \newrefformat{fig}{Fig.~\ref{#1}}

\title{Gigamachine: incremental machine learning on desktop computers}
\author{Eray {\"O}zkural \\
  Bilkent University \\ Ankara, Turkey }
\pubnote{\em Submitted To AGI-10} 
\begin{document}
\maketitle

\begin{abstract}
  We present a concrete design for Solomonoff's incremental machine
  learning system suitable for desktop computers. We use R5RS Scheme
  and its standard library with a few omissions as the reference
  machine. We introduce a Levin Search variant based on a stochastic
  Context Free Grammar together with new update algorithms that use
  the same grammar as a guiding probability distribution for
  incremental machine learning. The updates include adjusting
  production probabilities, re-using previous solutions, learning
  programming idioms and discovery of frequent subprograms. The issues
  of extending the a priori probability distribution and bootstrapping
  are discussed.  We have implemented a good portion of the proposed
  algorithms.  Experiments with toy problems show that the update
  algorithms work as expected.
\end{abstract}

\section{Introduction}

Artificial General Intelligence (AGI) field has received considerable
attention from researchers in the last decade, as the computing
capacity marches towards human-scale. Many interesting theoretical
proposals have been put forward
\cite{solomonoff-incremental,Hutter:01fast,Schmidhuber:09gm} and
practical general-purpose programs have been demonstrated (for
instance \cite{oops,rudi-clustering-compression}). We currently
understand the requirements of such a system much better than we used
to, therefore we believe that it is now time to start constructing an
AGI system, or at least a prototype based on the solid theoretical
foundation that exists today. We anticipate that during the tedious
work of writing such general purpose AI programs, we will have to
solve various theoretical problems and deal with practical details. It
would be for the best if we expose those problems early on.

Gigamachine is our initial implementation of an AGI system in the
O'Caml language with the goal of building what Solomonoff calls a
``Phase 1 machine'' that he plans to use as the basis of a quite
powerful incremental machine learning system
\cite{solomonoff-progress}.  While a lot of work remains to implement
the full system, the present algorithms and implementation demonstrate
a lot of issues in building a realistic system. Thus, we report on our
ongoing research to share our experience in designing such a system.
Due to space restrictions we cannot give much background, and we 
proceed directly to our contributions. 

\section{Scheme as the reference machine}

\cite{solomonoff-theoryandapps} argues that the choice of a reference
machine introduces a necessary bias to the learning system and looking
for the ``ultimate machine'' may be a red herring. In \cite{oops}, the
program-size efficient FORTH language was employed to great effect. In
\cite{solomonoff-progress} and other publications of Solomonoff, we
see that an as of yet (seemingly) unimplemented reference machine called AZ is
introduced. The AZ language is a functional programming language that
adopts a prefix (Polish) notation for expressions.
 Solomonoff also suggests adding primitives such
as $+$, $-$, $*$, $/$, $\sin$, $\cos$, etc. For a specific application,
one must choose a universal computer with as many suitable primitives
as possible, for it would take a lot of time for the system to
discover those primitives on its own, and the training sequence would
have to be longer to accommodate for the discovery of those primitives.

For a general purpose machine learning system, we need a general
purpose programming system that can deal with a large variety of data
structures and makes it possible to write sophisticated programs of
any kind. While FORTH has yielded rather impressive results, we have
chosen R5RS Scheme on the grounds that it is a simple yet general
purpose high-level programming language. Certain other features of it also make
it desirable.  Scheme was invented by Guy Lewis Steele Jr. and Gerald
Jay Sussman \cite{scheme}.  It is an improvement over LISP in that it
is statically scoped and its implementations are required to have
proper tail recursion; R5RS Scheme is defined precisely in a standards
document \cite{r5rs}. R5RS contains a reasonably sized standard
library. We do not think that Scheme has any major handicaps compared
to AZ. The small syntactic differences are not very important, but
language features are.  Scheme does include a functional language, in
addition to imperative features. It is highly orthogonal as it is
built around symbolic expressions. The syntax-semantics mapping is
quite regular, hence detecting patterns in syntax helps detecting
patterns in semantics.  There are a lot of efficient interpreters for
Scheme, which may be modified easily for our uses (we used the ocs
interpreter with a Scheme execution cycle limit that we added). 
Static scopes mean that the variable access is fast and
uncomplicated.  Scheme R5RS is quite high level, it has all the basic
data structures like lists, vectors, and strings. It can work with a
variety of numbers like integers, rationals, reals (also unlimited
precision reals), complex numbers and the standard library contains
the mathematical functions associated with these number types.

\subsection{Adaptation to program generation}

While we do not think that Scheme R5RS is the ultimate reference
machine, it has formed a good platform for testing out our ideas about
incremental learning. We have implemented most of the R5RS syntax,
with a few omissions. We have elected to exclude the syntax for
quasi-quotations and syntax transformation syntax, as the advanced
macro syntax would complicate our grammar based guidance logic, and as
it is an advanced feature that is used only in more complex programs,
which we do not expect to generate in the Gigamachine.  Further
simplifications were deemed necessary. In some parts of the syntax,
the same semantics can be expressed in different ways, for instance,
\texttt{(quote (0 1))} and \texttt{`(0 1)} have the same semantics. In
that case, we only used \texttt{quote}. In the case of number
literals, we do not generate alternative radii and use only base 10.
All of the R5RS standard library has been implemented except for
input/output (6.6) and system interface (6.6.4) forming an adequate
basis for generating simple programs. A special non-terminal called
\nt{standard-procedure} was added to the grammar which produces
standard library procedure calls, with the correct number of
arguments. The \nt{standard-procedure} is added as an alternative
production of the \nt{procedure-call} head in the Scheme standard grammar.
Further useful libraries common across Scheme implementations may be
easily added to the present system.

\section{Program Search}

In many AGI systems, a variant or extension of Levin Search
\cite{levin-search} is used for finding solutions. Solomonoff's
incremental machine learning also uses Levin Search as the basic
search algorithm to find solutions \cite{solomonoff-incremental}. In
our system, we take advantage of the stochastic grammar based guiding
probability mass function (pmf) for the search procedure as well. 
We first describe a generalized version of Levin Search and then build on it.

\subsection{Generalized Levin Search}

In \prettyref{alg:lsearch} we give a generalized Levin Search
algorithm similar to the one described in \cite{oops}. Inputs are as
follows. $U$ is a universal computer and $U(p,t)$ executes a program p
up to a duration of $t$ and returns its output.  $G$ is a grammar that
defines the language of valid programs (and $\fancy{L}(G)$ is its
language) in $U$.  $P$ is an a priori pmf 
of programs of $U$.  \algo{TestProg} is an algorithm that takes a
candidate program $x$ and forms a test program in the program coding
of $U$.  \var{TrueVal} is the value of ``true'' in the language of
$U$.

The constant $t_0$ is the initial time limit and the constant $t_q$ is
the time quantum that is the minimum time we run a program.

We start with a global time limit $t$ equal to $t_0$. We then start an
infinite loop.  Within an iteration, we allocate time to all the
programs in proportion to their a priori probabilities.  We choose a
set of candidate programs $C$ among the language of $G$ such that the
allocated time of a program is greater than or equal to $t_q$.  For
each program $c$ in $C$, we construct a test program using the
algorithm $\algo{TestProg}$. We execute the test program up to the
time limit $P(c).t$. Thus, the total time of running and testing
candidate programs do not exceed $t$ (with some work, the cost of
generation can be added as well). If the test is successful, the
search procedure returns $c$. Otherwise, after all the candidate
programs are tested in the iteration, the time limit is doubled, and
the search continues.

\begin{algorithm}
  \begin{small}
  \caption{$\algo{LSearch}(U, G, P, \algo{TestProg}, \var{TrueVal})$}
  \label{alg:lsearch}
  \begin{algorithmic}[1]
    \STATE $t \gets t_0$ \WHILE{ true } \STATE $C \gets \{ x \in
    \fancy{L}(G) ~|~ P(x).t \geq t_q \}$ \FORALL{$c \in C$} \IF{
      $U(\algo{TestProg(c)}, P(c).t) = \var{TrueVal}$} \RETURN $c$
    \ENDIF
    \ENDFOR
    \STATE $t \gets 2t$
    \ENDWHILE
  \end{algorithmic}
  \end{small}
\end{algorithm}

\vspace*{-16pt}
\subsection{Using a stochastic CFG in Levin Search}

A stochastic CFG is a CFG augmented by a probability value on each
production.  For each head non-terminal, the probabilities of
productions of that head must sum to one, obviously.

We can extend our Levin Search procedure to work with a stochastic CFG
that assigns probabilities to each sentence in the language. For this,
we need two things, first a generation logic for individual sentences,
and second a search strategy to enumerate the sentences that meet the
condition in Line 3 of \prettyref{alg:lsearch}.

In the present system, we use leftmost derivation to generate a
sentence, intermediate steps are thus left-sentential forms
\cite[Chapter 5]{automata}. The calculation of the a priori
probability of a sentence depends on the obvious fact that in a
derivation $S \Rightarrow \alpha_1 \Rightarrow \alpha_2 \Rightarrow
...  \Rightarrow \alpha_n$ where productions $p_1, p_2, ..., p_n$ have
been applied in order to start symbol $S$, the probability of the
sentence $\alpha_n$ is naturally $P(\alpha_n) = \prod_{1\leq i \leq n}
p_i$. Note that the productions in a derivation are conditionally
independent. While this makes it much easier for us to calculate
probabilities of sentential forms, it limits the expressive power of
the probability distribution.

A relevant optimization here is starting not from the absolute start
symbol (in the case of R5RS Scheme \nt{program}) but from any
arbitrary sentential form. This helps fixing known parts of the
program that is searched, and we have done so in the implementation.

The search strategy is important for efficient and correct
implementation of the for loop in Line 4 of \prettyref{alg:lsearch}.
We examine two relevant search strategies.

\subsection{Depth-limited depth-first search}

Depth-first search is a common search strategy when the search space
is large and the depth is manageable. We make use of a depth-limit in
the form of a ``probability horizon'' which is a threshold we impose
corresponding to the smallest probability sentence that we are willing
to generate. The probability horizon can be calculated from $t$ and
$t_q$ as $p_h = t_q/t$, which ensures that we will not waste time
generating any programs that we will not run.  The depth-first search
is implemented by expanding the leftmost non-terminal in a sentential
form, pruning the sentential forms which have a priori probabilities
smaller than the probability horizon, sorting the expanded sentential
forms in the order of decreasing a priori probability and then
recursively searching the list of sentences obtained in that fashion.
The recursion can be implemented via a stack or plain recursion. Stack
implementation turned out to be a bit faster.

\subsection{Best-first search and hybrid search}

A problem with the depth-first search order is that the search
procedure can terminate with a program that has a smaller a priori
probability than the best solution in current $C$. To amend this
shortcoming, we have tried a best-first search strategy, which
maintains a global priority queue during the expansion of the leftmost
non-terminal. Since every non-terminal has to be eventually expanded
this should indeed maintain a global order. However, maintaining
best-first search has a high memory cost, therefore in the
implementation we disabled this strategy as it quickly exhausts
available memory.

A solution that we have not yet tried is a hybrid search strategy.
Hybrid search was first proposed in \cite{oops-tr} as 50\%-50\% time sharing
of breadth-first and depth-first search for a simpler program probability
model. When using stochastic CFG's, there may be many non-terminals to expand in a
sentential form, therefore strict breadth-first search may incur a
very high memory cost. Note that by lazily expanding nodes, we can fix
this memory cost problem. The hybrid search strategy however, can be
used to run either breadth-first or best-first alongside depth-first
search. An improved hybrid search strategy is memory aware; it runs
best-first or breadth-first until the search queue structure reaches a
certain size, and then switches dynamically to depth-first search.

Solomonoff also suggests a randomized \algo{LSearch} which we did not
try as the disk swapping scheme seems too expensive
\cite{solomonoff-progress}.

\subsection{Generation of literals}

In \cite{solomonoff-theoryandapps} the Rissanen distribution $P(n) =
A2^{-log_2^*n}$ is proposed for generation of integer literals. An
alternative is the Zeta distribution with the pmf given by
\begin{equation*}
  \label{eq:zeta}
  P_s(k) = k^{-s} / \zeta(s)
\end{equation*}
where $\zeta(s)$ is the Riemann zeta function.  We have used the Zeta
distribution with $s=2$ and used a pre-computed table to generate up
to a fixed integer (1024 in our current implementation).  The Zeta
distribution has empirical support, that is why it was preferred.  The
upper bound is present to avoid too many programs with equal
constants. The expression syntax handles larger integers, for instance
\code{(* 1024 200)}. A smaller or greater upper bound may be
appropriate, this is a matter of experimentation.

The string literals are generated as a sequence of characters in the
grammar, as the default sequence rules seemed reasonable. We did not
see the need to come up with a specialized generation, but of course a
Zipf-distribution may be appropriate for that purpose.

To implement these special distributions and other kinds of rules, we
have defined a second kind of non-terminal which we call a non-terminal
procedure. A non-terminal procedure in our implementation is a function
that generates a list of sentential forms and associated probabilities
that can be used by the sequential enumeration algorithm. In the
below, it will also be used to implement a special kind of
context-sensitivity.

\subsection{Defining and referencing variables}

A major problem in our implementation was the abundance of ``unbound
reference'' errors. We have devised a simple solution that can be
implemented easily.

We maintain a static environment during leftmost derivation of a
sentence that is passed along to the right, possibly with
modification. The environment is also passed along to non-terminal
procedures. Initially, the environment includes the input parameters
to the function that is being searched. When a variable is defined, a
robotic variable name is generated in the form of
\code{var}\nt{integer} where the non-terminal \nt{integer} is sampled
from the Zeta distribution. We use the same pre-computed Zeta
distribution and the same upper bound as in integer literal
generation. This is not a drastic limitation as it limits only
temporary variables within the same scope.  When a variable reference
is generated, the environment is present and we choose among available
variable names with uniform distribution.

Nevertheless, this fix does not respect the nesting of scopes.
If we are expanding a \nt{definition} (as in \code{(define a 3)}, then
the variables defined in that definition must be available in the
enclosing scope.  This requires the generator to be aware of scope
beginning and ending. Within a scope, the definitions, excluding
mutually recursive definitions, are available from the point of
definition until the scope end. Robotic variable definitions are seen
in \code{define}, \code{lambda}, \code{let}, \code{let*},
\code{letrec} and \code{do} blocks. Mutually recursive
  definitions in \code{letrec} blocks have to be handled in a
  different way by first generating all the robotic variables and then
  generating the rest of the bindings.


The current implementation does not have any such scope begin/end
awareness since it is unlikely to generate very long programs.  It
propagates environment modifications from left to right disregarding
any nesting of scopes. To make it aware without rewriting everything,
we might try to annotate the rules that define new scopes with special
non-terminal symbols \nt{scope-begin} and \nt{scope-end}.  Instead of
passing a single environment to the partial derivation, we can pass a
stack of environments. when a \nt{scope-begin} is seen, the current
stack is pushed, and when a \nt{scope-end} is seen the stack is
popped.


\section{Stochastic CFG updates}

The most critical part of our design is updating the stochastic CFG so
that the discovered solutions in a training sequence will be more
probable when searching for subsequent problems. We propose four
update algorithms that work in tandem.

\cite{solomonoff-theoryandapps} mentions PPM \cite{cleary-adaptive}.
PPM can indeed be used to extrapolate a set of programs, however we do
not think it is practical for incremental machine learning. In fact,
we had adopted one of the recent variants of PAQ family of compression
programs \cite{mahoney-adaptive} for this purpose (by first
compressing the set of programs and then appending bits to the end of
the stream dynamically during decompression), and we saw that the
extrapolated programs were mostly syntactically incorrect. We can devise a PPM
variant that is useful for extrapolating programs but this would not
make our job easier. The update algorithms that we propose are more
powerful.

\subsection{Modifying production probabilities}

The simplest kind of update is modifying the probabilities as new
solutions are added to the solution corpus. For this, however, the
search algorithm must supply the derivation that led to the solution
(which we do), or the solution must be parsed using the same grammar.
Then, the probability for each production $A \rightarrow \beta$ in the
solution corpus can be easily calculated by the ratio of frequency of
productions $A \rightarrow \beta$ in the solution corpus to the
frequency of productions in the corpus with a head of $A$. The
non-terminal procedures are naturally excluded from the update as they
can be variant. However, we cannot simply write the probabilities
calculated this way over the initial probabilities, as initially there
will be few solutions, and most probabilities will be zero. We use
exponential smoothing to solve this problem.
\begin{eqnarray*}
  \label{eq:expsmoothing}
  s_0 &=& p_0 \\
  s_t &=& \alpha p_t + (1-\alpha) s_{t-1}
\end{eqnarray*}
where $p_0$ is the initial probability, $p_t$ is the probability in
the corpus and $\alpha$ is the smoothing factor. We used a smoothing
factor of $0.2$.  See
\cite{Merialdo93taggingenglish} for the application of smoothing in a
similar problem. Other methods like Laplace's rule may be used to 
avoid zero probabilities \cite{solomonoff-progress}. 

While modifying production probabilities is a useful idea, it cannot
add much information to the guiding pmf as the total amount of
information is limited by the number of bits per probability
multiplied by the number of probabilities. While we can use arbitrary
precision floating point numbers, it does not seem likely that
distinguishing more finely among a few number of alternative
productions for a non-terminal will result in great improvements. Then,
it seems that we need to augment the grammar with new productions. An
idea we have thought but not yet tried is to convert occurrences of the
same non-terminal into multiple non-terminals, so they will have
different probabilities as a result of learning. A collection of
non-terminals can be replicated in this way, as well. However, of
course, this replication is also limited and does not seem to overcome
the structural limitation of modifying probabilities.

\subsection{Re-using previous solutions}

In the course of a training sequence, the solutions can be
incorporated in full by adding the solutions to the grammar. In the
case of Scheme, there could be many possible implementations. The
simplest way we have found is to add all the solutions to the library
of the Scheme interpreter, add a hook non-terminal
\nt{previous-solution} to the grammar, and then extend the
\nt{previous-procedure} with the syntax to call the new solution. We
assume that this syntax is provided in the problem definition. We add
new solutions as follows, the new solution among other previous
solutions is given a probability of $\gamma$ in the hope that this
solution will be re-used soon, and then the probabilities of the old
productions of \nt{previous-solution} are normalized so that they sum
to $1-\gamma$. We currently use a $\gamma$ of $0.5$.

If it is impossible or difficult to add the solutions to the Scheme
interpreter as in our case, then all the solutions can be added as
\code{define} blocks in the beginning of the program produced. The
R5RS Scheme, being an orthogonal language, will allow us to make
definitions almost anywhere. However, there will be a time penalty
when too many solutions are incorporated, as they will have to be
repeatedly parsed by the interpreter during \algo{LSearch}. To solve
this problem and make the search a bit more scalable, we add a
hook called \nt{solution-corpus} to the grammar for \nt{definition},
which can be achieved in a similar way to \nt{previous-solution}.
However, then, the probability of defining \emph{and} using a previous
solution will greatly decrease. Assume that a previous solution
is defined with a probability of $p_1$ and called with a
probability of $p_2$. Since the grammar does not condition calling a
previous solution on the basis of definition, the probability of a
correct use is $p_1.p_2$; most of the time this logic will just
generate semantically incorrect invocations of the past solutions. To fix
this undesirable situation, we can use a non-terminal procedure in the
definition production for the particular solution, that stores the
solution in the environment that already stores 
variable names, defines the solution name as a variable, 
and in the production that calls the previous
solution, selects among those solutions in the environment with uniform
probability. When the solution is present we return a 
nil production with $0$ probability to avoid generation of 
redundant programs.

Since this is a complex solution, for other implementers it may be
preferable to just add the new solutions to the interpreter in a format 
that can be executed efficiently. Many Scheme interpreters have such 
``compiled'' data structures that the interpreter first converts to 
after parsing the program, and the evaluator is designed to 
work on those structures.

\subsection{Learning programming idioms}

Programmers do not only learn of concrete solutions to problems, but
they also learn abstract programs, or program schemas. One way to
formalize this is that they learn sentential forms. If we can extract
appropriate sentential forms, we can add these to the grammar via the
same algorithm that modifies production probabilities.

We have not yet implemented this update algorithm because we use a
leftmost derivation which does not immediately contain appropriate
sentential forms. Appropriate sentential forms can be obtained by
deriving from a relevant start symbol like \nt{body} (which is the
body of a Scheme function definition) or \nt{expression} (which is the
basic symbolic expression syntax that is used many times in the Scheme
grammar) until a particular \emph{level} or a \emph{border} in the
derivation tree of a solution body or a top-level expression. Thus,
some sub-expressions will remain unexpanded. To implement this in a
leftmost derivation scheme as ours, we may either parse the found
solutions, or change the derivation logic so that a derivation tree is
constructed during search. It is also possible to construct the
derivation tree from the leftmost derivation.

It is hoped that the system will be able to learn programming idioms
like that of recursion patterns, or ways to use loops via this kind of
update. For instance, assume that it discovered a kind of integer
recursion for a problem: \code{(define (myrec n) (if (= n 1) 0 (+ 1
  (myrec (/ n 2)))))}. Then, the sentential forms at intermediate
levels of the derivation tree for the \emph{body} of the solution
would abstract some of the sub-expressions, resulting for instance in
\code{(if (= \nt{variable} \nt{uinteger-10}) \nt{uinteger-10} (+
  \nt{uinteger-10} (\nt{variable} \nt{variable} \nt{uinteger-10})))}
if we pruned one level up from each leaf of the derivation tree (since
the derivation trees are usually quite unbalanced) as determined from
a hand made derivation tree, which may come in handy when dealing with
similar recursions. Several sentential forms can be learnt from a
single solution in this fashion corresponding to different syntactic
abstractions, and the algorithm for re-using previous solutions can be
invoked to add them to the grammar. Otherwise, we would have to teach
that kind of recursion through higher-order functions (which is also
an admissible strategy).

\subsection{Frequent sub-program mining}

Mining the solution corpus would enhance the guiding probability
distribution.  Frequent sub-programs in the solution corpus, i.e.,
sub-programs that occur with a frequency above a given
support threshold, can be added again as alternative productions to
the commonly occurring non-terminal \nt{expression} in the Scheme
grammar.  For instance, if the solution corpus contains several
\code{(lambda (x y) (* x y) )} subprograms the frequent sub-program
mining would discover that and we can add it as an alternative
expression to the Scheme grammar. We have not yet detailed this update
algorithm but it seems reasonable and it can benefit from the
well-developed field of data mining.

\section{Discussion}

\subsection{What does a programmer know?}

In order to encode useful information in the a priori probability
distribution, we must reflect on what a human programmer knows when
she writes a program.  The richness of the R5RS Scheme language
requires us to solve some problems like avoiding unbound references to
make even the simplest program searches feasible, therefore it is
important that we encode as much a priori knowledge about programming
as possible into the system. Among other things, a programmer knows
the following.  \emph{a)} The syntax of the programming language,
sometimes imperfectly. Our system knows the syntax perfectly, and does
not make any syntactic mistakes.  \emph{b)} The semantics of the
programming language, again imperfectly. Our system knows little about
writing semantically correct programs and often generates incorrect
programs. We may need to add more semantic checks to enhance that. Our
system does not know the referential semantics of the programs, only
how to run the program. It may be useful for the search procedure to
be informed of more semantics. Human programmers use semantic
information to accelerate writing programs, for instance by using
proper types.  \emph{c)} The running time and space complexity of the
program, imperfectly. The more computer science a programmer knows,
the fewer mistakes she will make. It can be argued that the
programmers have partial knowledge of the halting problem, as they
know many programs which will loop infinitely, and avoid writing them.
They also learn which programs \emph{seem} to loop indefinitely.
\emph{d)} Pre and post conditions. Sometimes when the programs are
well-specified, the programmer understands the conditions that will be
assumed prior to the running of the program and after the program
finishes.

Some of this information can be incorporated in the probabilistic
model, for others we may need to augment our system with relevant
algorithms from fields like automated theorem proving.

\subsection{Bootstrapping problems}

The incremental machine learning capabilities of the Phase 1 machine
in \cite{solomonoff-progress} will be used to calculate the
conditional distributions that are necessary for the Phase 2 machine
to take off. Our current implementation can find short programs, but
despite any future improvements to processing speed, it may have
difficulty in finding the required sort of programs without a huge
training sequence that approaches those programs very closely, and it
has little chance of rewriting itself unless the current
implementation is further developed.


\section{Experiments}

Currently, we have made our implementation work only on toy problems.
We initially solved problems for inverting mathematical functions, for
identity function, division, and square-root functions to test
\algo{LSearch}. We then developed a simple training sequence composed
of a series of problems. For each problem, we have a sequence of input
and output pairs, and we use incremental operator induction
\cite{solomonoff-progress,solomonoff-threekinds}. The operator
induction, in a similar way to OOPS \cite{oops} first finds a solution
to the first pair, then, the first and second pairs, and then the
first three, and so forth. After each partial or complete solution to
a problem, a stochastic CFG update is applied.  In our first toy
training sequence, we have the identity function \code{id}, the square function \code{sqr},
the addition of two variables \code{add}, a function to test if the argument is
zero \code{is0}, all of which have $3$ example pairs, 
fourth power of a number \code{pow4} with just $2$ example pairs, boolean 
\code{nand}, \code{nor} and \code{xor} functions with $4$ example pairs each, 
and the \code{factorial} function $f(x)=x!$ with $7$
pairs for inputs $0\ldots6$. The \code{factorial} function took more
than a day at the fifth example ($f(4)=24$) so we interrupted it
as it is not feasible on a desktop machine, it did have a partial solution from the first four pairs.  However, we have observed
that the two update algorithms implemented work gracefully. The search
time for the later problems tend to reduce compared to the first.
Since we extend the grammar, sometimes a slight slowdown is experienced,
which would be amortized in later problems depending on the training sequence.
 For instance,  
the \code{sqr} problem took $25806$ trials while the \code{add} problem took $32222$. 
The solution of logical functions took 
longer than previous problems, but eventually we saw time reductions in 
them. The \code{nand} solution \code{(not (if y x y))} took $8413333$ trials, 
the next problem nor took $427582$ trials, and about $30$ speedup in Scheme cycles. 
Previous solutions are re-used aggressively. \code{sqr} problem is solved
in the first example of $f(6)=36$ and incorporating the next example takes
only $210$ trials corresponding to about $100$ speedup. \code{pow4} 
solution \code{(define (pow4 x ) (define (sqr x )   (* x
x))  (sqr (sqr x ) ))} re-uses the \code{sqr} solution and takes $25056$ trials, faster than solving \code{sqr} itself.
The update algorithms help the machine learn something about
semantics. While searching for \code{sqr}, at seventh
iteration, the system reported $97\%$ error rate in evaluation of
candidate programs. In the last problem, the error rate dropped to
$88\%$ in the same iteration, although we did not consider any 
semantics in the update algorithms. Error reductions are seen also 
when code re-use is disabled.

\section{Conclusion}

We have described a stochastic CFG based incremental machine learning
system targeting desktop computers in detail. We have adapted R5RS
Scheme as the reference universal computer to our system. The
stochastic CFG is used in sequential \algo{LSearch} to calculate a
priori probabilities and to generate programs efficiently avoiding
syntactically incorrect programs. We derive sentences using leftmost
derivation. We use a probability horizon to limit the depth of
depth-first search, and we also propose using best-first search and
memory-aware hybrid search. We have specialized productions for number
literals, variable bindings and variable references; in particular, we
avoid unbound variables in generated programs. We have proposed four
update algorithms for incremental machine learning. Two of them have
been implemented. To the extent that the update algorithms work, their
use has been demonstrated in a toy training sequence.

The slowness of searching the factorial function made us
realize that we need major improvements in both the search and the
update algorithms if we would like to continue using Scheme. 
We have been working on a more realistic training 
sequence that features recursive problems, optimizing search and 
implementing remaining update algoritms. 
After that, we will extend our implementation to
work on parallel hardware, implement the Phase 2 of Solomonoff's
system, and attempt incorporating features from other AGI proposals such as
\algo{HSearch} and G{\"o}del Machine. We would like to implement
alternative approaches and compare them, as well.

\bibliography{agi}

\begin{thebibliography}{Mah05}

\bibitem[GJS75]{scheme}
Jr. Gerald Jay~Sussman, Guy Lewis~Steele.
\newblock Scheme: An interpreter for extended lambda calculus.
\newblock Technical Report AI Lab Memo AIM-349, MIT AI Lab, December 1975.

\bibitem[Hut02]{Hutter:01fast}
Marcus Hutter.
\newblock The fastest and shortest algorithm for all well-defined problems.
\newblock {\em International Journal of Foundations of Computer Science},
  13(3):431--443, June 2002.

\bibitem[JC84]{cleary-adaptive}
I.H.~Witten J.G.~Cleary.
\newblock Data compression using adaptive coding and partial string matching.
\newblock {\em IEEE Transactions on Communications}, 32(4):396--402, April
  1984.

\bibitem[JEH01]{automata}
JeffreyD.~Ullman John E.~Hopcroft, Rajeev~Motwani.
\newblock {\em Introduction to Automata Theory, Languages, and Computation}.
\newblock Addison Wesley, second edition, 2001.

\bibitem[Lev73]{levin-search}
L.~A. Levin.
\newblock Universal sequential search problems.
\newblock {\em Problems of Information Transmission}, 9(3):265--266, 1973.

\bibitem[Mah05]{mahoney-adaptive}
Matt Mahoney.
\newblock Adaptive weighing of context models for lossless data compression.
\newblock Technical Report CS-2005-16, Florida Tech., 2005.
\newblock Describes paq6.

\bibitem[Mer93]{Merialdo93taggingenglish}
Bernard Merialdo.
\newblock Tagging english text with a probabilistic model.
\newblock {\em Computational Linguistics}, 20:155--171, 1993.

\bibitem[RC03]{rudi-clustering-compression}
P.M.B.~Vitanyi R.~Cilibrasi.
\newblock Clustering by compression.
\newblock Technical report, CWI, 2003.

\bibitem[RK98]{r5rs}
Jonathan~Rees Richard~Kelsey, William~Clinger.
\newblock Revised5 report on the algorithmic language scheme.
\newblock {\em Higher-Order and Symbolic Computation}, 11(1), 1998.
\newblock (editors).

\bibitem[Sch02]{oops-tr}
Juergen Schmidhuber.
\newblock Optimal ordered problem solver.
\newblock Technical Report TR IDSIA-12-02, IDSIA, 2002.
\newblock Revision 2.

\bibitem[Sch04]{oops}
Juergen Schmidhuber.
\newblock Optimal ordered problem solver.
\newblock {\em Machine Learning}, 54:211--256, 2004.

\bibitem[Sch09]{Schmidhuber:09gm}
J.~Schmidhuber.
\newblock Ultimate cognition {\em \`{a} la} {G\"{o}del}.
\newblock {\em Cognitive Computation}, 1(2):177--193, 2009.

\bibitem[Sol89]{solomonoff-incremental}
Ray Solomonoff.
\newblock A system for incremental learning based on algorithmic probability.
\newblock In {\em Proceedings of the Sixth Israeli Conference on Artificial
  Intelligence}, pages 515--527, Tel Aviv, Israel, December 1989.

\bibitem[Sol02]{solomonoff-progress}
Ray Solomonoff.
\newblock Progress in incremental machine learning.
\newblock In {\em NIPS Workshop on Universal Learning Algorithms and Optimal
  Search}, Whistler, B.C., Canada, December 2002.

\bibitem[Sol08]{solomonoff-threekinds}
Ray Solomonoff.
\newblock Three kinds of probabilistic induction: Universal distributions and
  convergence theorems.
\newblock {\em The Computer Journal}, 51(5):566--570, 2008.

\bibitem[Sol09]{solomonoff-theoryandapps}
Ray Solomonoff.
\newblock Algorithmic probability: Theory and applications.
\newblock In M.~Dehmer and F.~Emmert-Streib, editors, {\em Information Theory
  and Statistical Learning, Springer Science+Business Media}, pages 1--23.
  N.Y., 2009.

\end{thebibliography}
\bibliographystyle{alpha} 
\end{document}